\documentclass[11pt,a4paper]{article}
\usepackage[hyperref]{emnlp2020}
\usepackage{times}
\usepackage{amssymb}
\usepackage{latexsym}
\usepackage{graphicx}
\usepackage{caption}
\usepackage{booktabs}
\usepackage{array}
\usepackage{subcaption}
\usepackage{enumitem}
\usepackage{amsmath}

\usepackage{microtype}

\aclfinalcopy 
\def\aclpaperid{2182} 

\newcommand{\ptdata}{{\sc P2T-Rel}}
\newcommand{\qpdata}{{\sc Q2P-Mat}}
\newcommand{\berttiny}{{\sc BERT-Tiny}}
\newcommand{\decos}{{\sc DE-Cos}}
\newcommand{\deffnn}{{\sc DE-FFNN}}
\newcommand{\denmffnn}{{\sc DiPairFFNN}}
\newcommand{\denmtsf}{{\sc DiPairTSF}}

\title{DiPair: Fast and Accurate Distillation for Trillion-Scale \\ Text Matching and Pair Modeling}
       
\author{Jiecao Chen\Thanks{~Correspondence to chenjiecao@google.com}, Liu Yang, Karthik Raman, Michael Bendersky \\ \textbf{Jung-Jung Yeh, Yun Zhou, Marc Najork, Danyang Cai, Ehsan Emadzadeh}\\
  \{chenjiecao, yangliuy, karthikraman, bemike, jjyeh, yunzhou, najork, dycai, eemadzadeh\}@google.com\\
Google Research}

\date{}

\begin{document}
\maketitle
\begin{abstract}
Pre-trained models like BERT \cite{devlin2018:bert} have dominated NLP / IR applications such as single sentence classification, text pair classification, and question answering. However, deploying these models in real systems is highly non-trivial due to their exorbitant computational costs. 
A common remedy to this is knowledge distillation \cite{DBLP:journals/corr/HintonVD15}, leading to faster inference.
However -- as we show here -- existing works are not optimized for dealing with pairs (or tuples) of texts.
Consequently, they are either not scalable or demonstrate subpar performance.
In this work, we propose \textbf{DiPair} --- a novel framework for distilling fast and accurate models on text pair tasks. Coupled with an end-to-end training strategy, DiPair is both highly scalable and offers improved quality-speed tradeoffs.
Empirical studies conducted on both academic and real-world e-commerce benchmarks demonstrate the efficacy of the proposed approach with speedups of over 350x and minimal quality drop relative to the cross-attention teacher BERT model.
\end{abstract}

\section{Introduction} 

Modeling the relationship between textual objects is critical to numerous NLP and information retrieval (IR) applications \cite{10.5555/2683840}. This subsumes a number of different problems such as textual entailment, semantic text matching, paraphrase identification, plagiarism detection, and relevance modeling. For example, modeling the relationship between queries and documents / ad keywords is central to search engines / digital advertisement systems \cite{10.5555/2683840,DBLP:journals/corr/abs-1903-06902}.


Recently, neural network-based models have demonstrated large gains in this space \cite{DBLP:conf/nips/HuLLC14,DBLP:conf/aaai/PangLGXWC16}. In particular, the Transformer / BERT family of models \cite{devlin2018:bert,albert19,roberta19,electra20} have set a new bar for these semantic text matching problems. However, the computational costs of these models have proven to be prohibitively expensive, thus limiting their use in real-world applications \cite{FC19}. For example, on the e-commerce relevance-scoring task (\ptdata~dataset) discussed in Sec. \ref{sec:datasets}, scoring the (trillion+) text pairs would take years.



One popular remedy is to distill these expensive \emph{teacher} models \cite{DBLP:journals/corr/HintonVD15} into lightweight \emph{student} models. Training these \emph{students} using examples labeled by the \emph{teacher} has been shown to maintain quality while enabling faster inference. The key to the effectiveness of distillation techniques is a good trade-off between student quality and inference speed. 

However, as we show here, existing knowledge distillation techniques \cite{DBLP:journals/corr/abs-1910-01108,DBLP:journals/corr/abs-1909-10351,DBLP:journals/corr/abs-1908-08962,DBLP:journals/corr/abs-1903-12136} fall short on the quality-speed trade-off when dealing with pairs of texts. On one hand, approaches that model the texts jointly (\emph{i.e.}, using cross-attention) even one as highly optimized as \berttiny~\cite{DBLP:journals/corr/abs-1908-08962} are still orders of magnitude too slow. 

On the other hand, techniques that model the texts independently such as the \textit{dual-encoder} models\footnote{These models encode the two texts separately and then combine them via a lightweight dot product / cosine.} \cite{das-etal-2016-together,JDH17,chidambaram-etal-2019-learning,cer-etal-2018-universal,DBLP:journals/corr/HendersonASSLGK17,reimers-gurevych-2019-sentence} are able to run efficient inference on large-scale text pairs.
By exploiting the independence of the texts, these techniques can significantly speed up inference by caching/indexing embeddings of individual texts.
However, this speedup comes at a significant cost -- with sharply reduced scoring quality.

The key drawback here is that these independent models lack the ability to mimic the cross-attention enabled teachers and model the joint nuances and facets of the texts.
As a motivating example, consider the ecommerce term relevance-scoring task. For the product ``Black Sport Nike Shoes for Boys Size Wide", terms such as ``black", ``wide footwear" and ``nike shoes" are all relevant. However, enforcing similarity between the independently modeled term and product will lead to the embeddings of ``black" and ``nike shoes" being incorrectly considered similar.





Motivated by this, we propose \textbf{DiPair} for fast and accurate distillation of large-scale text matching and pair modeling. DiPair aims to combine the best of both worlds: Like dual-encoder models, it leverages common pre-computation, while at the same time modeling the text jointly -- with cross-attention -- using multiple contextual embeddings for each text. In particular, we extract a small fraction of the output token embeddings from each text, and then jointly model this smaller ``sequence'' using a transformer \emph{head} (we use the term \emph{head} to refer to the component that consumes the outputs of a dual-encoder model, see Figure \ref{fig:di-pair-model}).
We demonstrate that a two-stage, end-to-end training allows the proposed DiPair model to learn richer multifaceted semantic representations of the text pairs. The resulting DiPair model is 350x+ faster with minimal quality drop relative to the teacher on academic and real-world e-commerce datasets. 

\begin{figure}[t]
    \centering
    \vspace{-0.1em}
    \includegraphics[width=0.8\columnwidth]{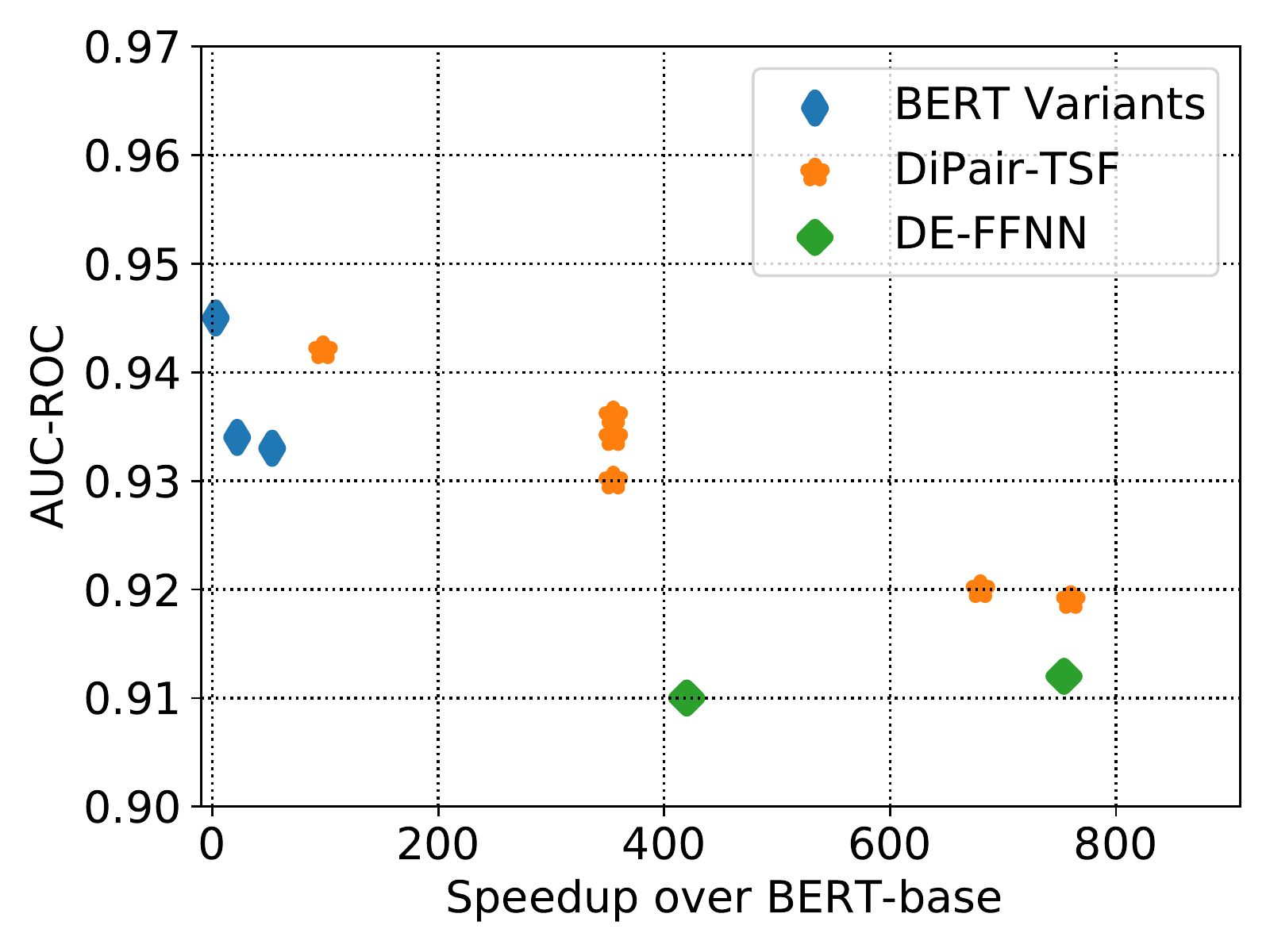}
    \vspace{-1em}
    \caption{Inference speed vs quality trade-off for three different representative approaches (see Sec. \ref{sec:baseline} for a more detailed description): DE-FFNN refers to dual-encoder + a feedforward neural network head; BERT variants refers to different cross-attention BERT-based student models released by \citet{DBLP:journals/corr/abs-1908-08962}; DiPair-TSF refers to the DiPair model with a transformer head. The metrics and inference speedup are evaluated with the \qpdata~dataset (see Sec. \ref{sec:datasets}). Note that the DiPair and DE-FFNN are varied using only the head in this plot for a fair comparison.} 
    
    \label{fig:dipair-tradeoff}
    \vspace{-.5em}
\end{figure}

In summary, our main contributions include:
\begin{itemize}
\itemsep0em
    \item \textbf{DiPair}: A new framework for distilling fast, accurate models on text pair tasks. Its advantages include: 1) Generic framework applicable across numerous applications involving pairwise/n-ary textual input. To the best of our knowledge, this is among the first few works tackling this problem. 
    2) Highly practical solution with limited storage and computation needs that scales to trillions of examples. 3) Large speedups for model inference -- 350x+ faster relative to the BERT-base teacher and 8x faster than previous highly optimized benchmarks \cite{DBLP:journals/corr/abs-1908-08962}. 
    \item A two-stage, end-to-end training scheme enables an  improved quality-speed tradeoff as shown in Fig.~\ref{fig:dipair-tradeoff}.
    \item Evidence that (self and cross) attention is important for student models when it comes to distilling from teachers like BERT.
    \item Extensive experiments on academic and real-world e-commerce datasets demonstrate that DiPair can lead to fast and accurate models that outperform existing techniques on text matching and pair modeling.
\end{itemize}

\section{Related Work}
\label{sec:related-work}
\textbf{Text Pair Modeling and Matching.}
A large variety of neural models have been proposed for text pair tasks such as matching and similarity scoring \cite{DBLP:conf/cikm/HuangHGDAH13,DBLP:conf/nips/HuLLC14,DBLP:conf/aaai/PangLGXWC16,Guo:2016:DRM:2983323.2983769,Yang:2016:ARS:2983323.2983818,Mitra:2017:LMU:3038912.3052579,DBLP:conf/sigir/XiongDCLP17,rao-etal-2019-bridging}. These models can be broadly classified into \textit{representation-focused} models (or \textit{dual-encoder} models) \cite{DBLP:conf/cikm/HuangHGDAH13,DBLP:conf/nips/HuLLC14} and \textit{interaction-focused} models \cite{DBLP:conf/aaai/PangLGXWC16,Guo:2016:DRM:2983323.2983769,Yang:2016:ARS:2983323.2983818,Mitra:2017:LMU:3038912.3052579,DBLP:conf/sigir/XiongDCLP17}, where the former involves encoding the individual text separately while the latter models the pair jointly (often involving some interaction / attention model). 
In recent years, Transformer \cite{NIPS2017_Transformers} based models like BERT \cite{devlin2018:bert} leveraged cross-attention to achieve impressive performance gains on several text pairs tasks including natural language inference \cite{bowman-etal-2015-large}, sentence pair classification and relevance scoring. As shown in several previous research \cite{DBLP:conf/aaai/PangLGXWC16,Guo:2016:DRM:2983323.2983769,Yang:2016:ARS:2983323.2983818,Mitra:2017:LMU:3038912.3052579,DBLP:conf/sigir/XiongDCLP17,devlin2018:bert}, interaction-focused models usually achieve better performances for text pair tasks. However, it is difficult to serve these types of models for applications involving large inference sets in practice. On the other hand, text embeddings from dual encoder models can be learned independently and thus pre-computed, leading to faster inference efficiency but at the cost of reduced quality. Early work like \cite{WJ17} uses attention to aggregate the two sequences of word embeddings, and a CNN model is then applied to extract the final representation. This method is relatively expensive as it requires to store the whole sequences of word embeddings and a full cross-attention operation has to be performed.   Recently the PreTTR model \cite{macavaney2020efficient} aimed to reduce the query-time latency of deep transformer networks by pre-computing part of the document term representations. However, their model still required modeling the full document/query input length in the head, thus limiting inference speedup. 
Another recent work is Poly-encoders \cite{HSL+20} which shared some similar motivations. However, Poly-Encoders makes strong assumptions on the input data property thus limiting its applicability (Appendix \ref{sec:d2d} demonstrates this quality drop on a standard text matching task). 



\textbf{Knowledge Distillation.} Our research is an example of knowledge distillation in neural networks \cite{DBLP:journals/corr/HintonVD15,DBLP:conf/emnlp/SunCGL19,DBLP:journals/corr/abs-1910-01108}. The idea of knowledge distillation is to transfer information from a heavily-parameterized and accurate teacher model to a lightweight student model for faster inference. \citet{DBLP:journals/corr/abs-1903-12136} proposed to distill knowledge from BERT to a single-layer BiLSTM model. TinyBERT \cite{DBLP:journals/corr/abs-1909-10351} performs knowledge distillation into transformers in two-stage learning including pre-training and task-specific fine-tuning. \citet{DBLP:journals/corr/abs-1908-08962} proposed Pre-trained Distillation, which shows task-specific distillation on an unlabeled transfer set is helpful to improve the student model performance. Key differences between our work and these approaches are that we focus on model distillation for text pair inputs and speeding up inference while aiming to match the teacher's performances.


\textbf{Model Quantization and Parameter Pruning.}
Another line of research loosely connected to our work is to reduce inference time via pruning less significant weights and/or converting the model to low-precision (aka quantization) \cite{HMD16,HZK+17,IMA+16,RFC20,FC19}. Effective in many applications, those approaches, however, often only lead to less than 20x speedup and therefore do not scale to many tasks with pairwise input.



\section{Our Approach}
\subsection{Method Overview}
\label{sec:method-overview}
\begin{figure*}[th]
	\center
	\includegraphics*[viewport=0mm 0mm 190mm 140mm, scale=0.80]{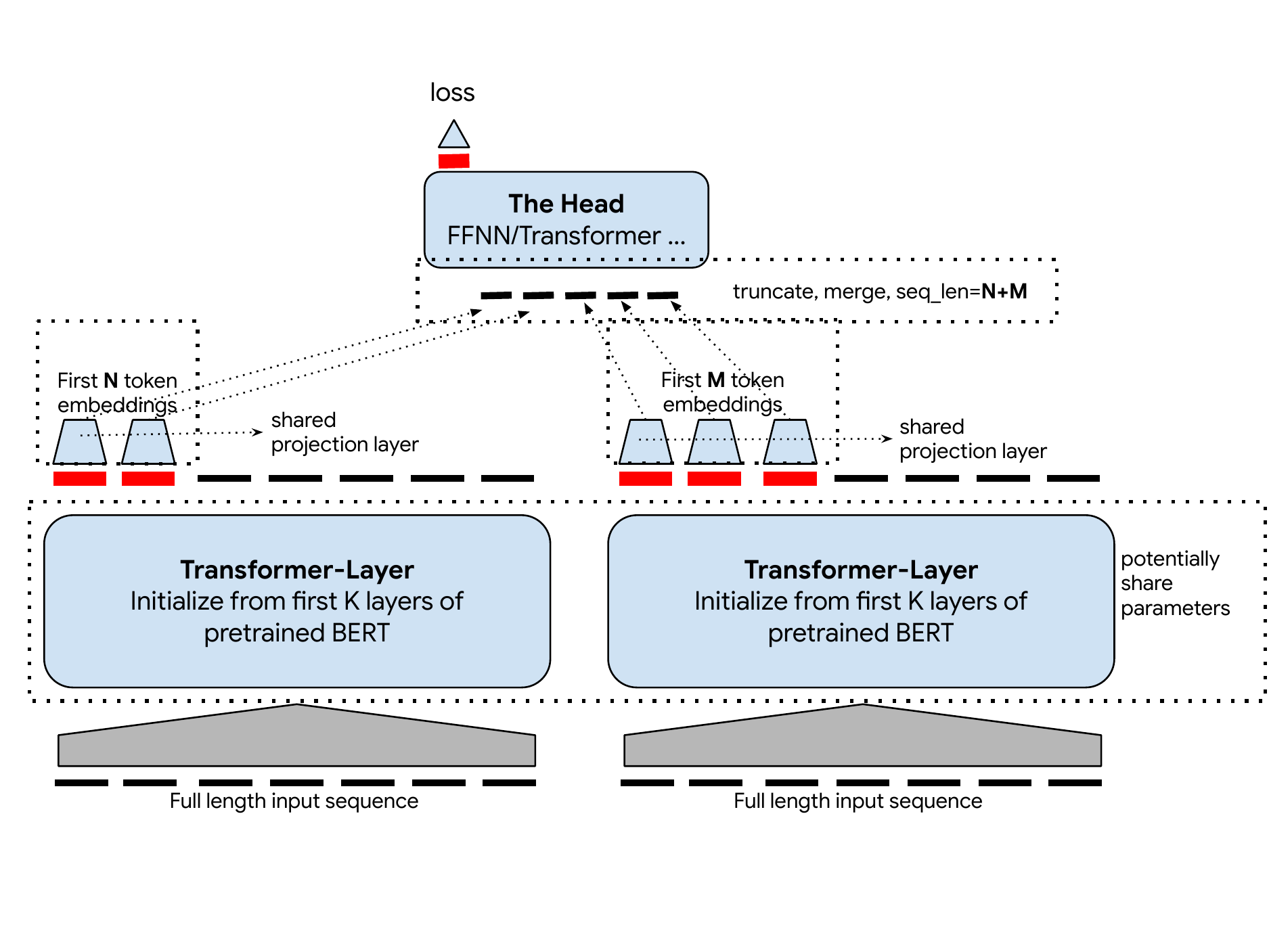}
	\vspace{-1.5cm}
	\caption{The architecture of the DiPair model.}\label{fig:di-pair-model}
	\vspace{-0.2cm}
\end{figure*}

Figure \ref{fig:di-pair-model} provides an overview of the proposed DiPair model. First, a transformer-based \emph{dual-encoder} model is applied to the input pair; the output of an encoder is a sequence of token embeddings, which has the same sequence length as the tokenized input text. We then \emph{truncate} the output sequences by only taking the first $N$ and $M$ token embeddings from the left and right inputs, respectively; the next step is to project those selected token embeddings into lower dimensions and merge them to form the new input sequence. The merged input sequence is then fed into the transformer (or an FFNN) \emph{head}, and the first token embedding of the output sequence of the head is used as the representation of the initial input pair. 

Note that, the dual-encoder will process the full-length input sequences. At the same time, the head only consumes a sequence of length $(N + M)$, which is typically much smaller than the length of the input sequences and ensures efficient execution of the head.

To create the training data for our proposed model, we use an expensive teacher model (e.g., a 12-layer BERT fine-tuned with human-rated data) to annotate a set of unlabeled text pairs (a.k.a. distillation set). The dual-encoder part of our model is initialized from the first few layers of a pre-trained BERT, and a novel \emph{two stage training strategy} (see Sec. \ref{sec:twostage}) is applied to boost the performance further. We defer more details of data specific model distillation to Sec. \ref{sec:model_distillation}.

We now discuss each component of the proposed architecture in detail.


\subsection{Dual-Encoder}
A dual-encoder is the key component of our proposed architecture, and we initialize our dual-encoder from pre-trained BERT (or tinyBERT, ALBERT, etc.). Our basic assumption is that the number of pairs is much larger than the set of unique inputs to the left or right encoders, and the bottleneck of serving our model is to run inference on the pairs with the head. Our proposed architecture, therefore, has an important benefit: 
increasing the model capacity does not increase the inference time as we can keep the head the same but use more expensive encoders.
Figure \ref{fig:encoder-layers} shows that increasing the number of layers of the encoders will often lead to better model performance.


\subsection{Truncated Output Sequences}
This is the key step to speed up the model serving. Recall that the running time of a transformer-based model quadratically depends on the input sequence length. One of the most effective ways to reduce the running time is to reduce the input sequence length. However, as Table \ref{fig:reduce-seq-len} reveals, blindly truncating the input to a BERT model will lead to a quick performance drop. Our key intuition is that, by using a dual-encoder + head architecture, we can focus on reducing the inference time of the head, instead of speeding up the encoders.

Therefore, we still use the full-length input sequences in our encoders, but aggressively reduce the input sequence length to the head. To be more concrete, before merging the outputted sequences from the two encoders, we take the first $N$ and $M$ token embeddings from the left and the right sequences, respectively; 
This truncation technique has several benefits:
\vspace{-.5em}
\begin{itemize}
    \itemsep0em 
    \item It significantly speeds up the inference with the head, as the time complexity of transformer layers is quadratic w.r.t. the input sequence length.
    \item It significantly reduces the amount of data we need to cache. Only the first few token embeddings need be stored as the output of the encoders.
    \item $N$ and $M$ can be tuned to reflect the desired effectiveness and efficiency trade-off for a particular problem domain.
\end{itemize}
\vspace{-.7em}
It is important to note that due to the end-to-end architecture of our model, even though we only use $(N + M)$ token embeddings from the output of the dual-encoder, the model learns to push the information of the input text to the first $(N + M)$ embeddings (thanks to the transformer layers, those selected token embeddings can interact with other token embeddings, and can be viewed as a summary of the full-length input sequences).

\subsection{Projection Layer}
For each encoder, we add a projection layer to project each token embedding to a lower dimension. A projection layer is shared within an encoder, but different encoders may use different projection layers. 
There are two purposes of adding the projection layers:
\vspace{-.5em}
\begin{itemize}
    \itemsep0em
    \item Reduce storage. To run the inference with the proposed architecture, we need to cache all the outputs from the encoders.
    \item Speed up the inference with the head. The time complexity of a transformer linearly depends on the embedding dimension.
\end{itemize}
\vspace{-.7em}

In Table \ref{tb:dim-projection}, we show that by choosing a proper projection layer, we can significantly reduce the embedding dimension with almost no quality drops.

\subsection{Transformer-Based Head}
After the projection layer, we merge the $N + M$ projected token embeddings into one sequence and feed it into the head. Like the BERT model, we also add position embeddings and segment embeddings to help the transformer head better aggregate the input sequence. The first token embedding (i.e., CLS embedding) of the transformer head is used as the final representation of the input pair.

Another advantage of using a head is that the head is \emph{tokenization-free}: the input to the head is purely float tensors, and we do not need to preprocess/ tokenize the input text. This may lead to an additional speedup.

It is worth mentioning that a feedforward neural network (FFNN) can also be used as a head. An FFNN is faster than a transformer-head and often gives reasonable performance (though worse than a transformer head). See the experimental section (Sec. \ref{sec:head}) for more discussion on these trade-offs.

\subsection{Task Specific Losses}
In the standard dual-encoder model and the recent Poly-Encoders \cite{HSL+20} work, the dot product between the embeddings is a scalar, which is not suited for tasks beyond regression/binary classification. On the other hand, our proposed architecture outputs a representation of the input pair and is therefore compatible with a wide range of loss functions.

\subsection{A Two-Stage Training Approach}
\label{sec:twostage}
It turns out that directly training the proposed models often leads to sub-optimal results (see Sec. \ref{exp:two-stage} for more evidence). This is primarily because adding non-trivial layers on top of a well pre-trained dual-encoder during training may corrupt the knowledge that has been preserved in the dual-encoder. To address this issue, we propose to use a \emph{two-stage training strategy}: we first freeze the dual-encoder part and only train the newly added parameters until convergence; we then unfreeze the dual-encoder and further train the entire model. A similar training strategy can be found in e.g., \cite{WSW+19}.


\subsection{Extension to $n$-Ary Tuple}
Unlike the models proposed in the recent works \cite{macavaney2020efficient,HSL+20} where only pairs can be supported, our proposed architecture trivially extends to the scenario where we have $n$-ary tuple of textual objects $(a_1, a_2, \ldots, a_n)$ as the model input, as we can simply replace the dual-encoder model with an $n$-encoder model. This feature is useful in many applications, such as QA tasks with context, or query to document scoring tasks with personalized information.


\section{Experiments}

In this section, we conduct experimental studies. We aim to answer the following questions through our experiments:
\vspace{-0.7em}
\begin{itemize}
\itemsep-0.3em
    \item RQ1: How well does our proposed architecture perform compared with other strong baseline approaches? Compared with the teacher, how much faster are our methods (Sec. \ref{sec:main_results})?
    \item RQ2: Compared with FFNN heads, is the transformer head essential to reduce the distillation gap (Sec.\ref{sec:head})?
    \item RQ3: How does two-stage training affect the final model performance (Sec. \ref{exp:two-stage})?
    \item RQ4: How would the proposed dual-encoder+head architecture be affected by other hyper-parameters of different components (Sec. \ref{sec:model-ablation}).
\end{itemize}

\subsection{Datasets}
\label{sec:datasets}
We evaluate our proposed methods on two datasets (Table \ref{tb:datasets} provides an overview):
\vspace{-0.5em}
\begin{itemize}
\itemsep-0.3em
    \item \qpdata~is a binary classification task derived from the MSMARCO Passage Ranking data\footnote{\url{https://microsoft.github.io/MSMARCO-Passage-Ranking/}}. Given a \emph{(query, passage)} pair, the goal is to predict whether the passage contains the answer for the query.  We measure the model performance using AUC-ROC.  Appendix \ref{sec:app:qpdata} lists more details.
    \item \ptdata~ is a regression task on a real-world ecommerce dataset. Given a \emph{(product, term)} pair, the goal is to predict the relevance of the term to the product. We measure the model performance using Pearson correlation with the  human judgments. Title and description are used as the product features. Appendix \ref{sec:app:ptdata} provides several examples of \emph{(product, term)} pairs.
\end{itemize}

\begin{table*}[!h]
   \centering
   \small
    \begin{tabular}{c|c|c|c|c|c|c|c|c}
    \hline \hline
    Data & \multicolumn{4}{c|}{\ptdata} & \multicolumn{4}{c}{\qpdata}\\
    \hline
    Item & Distill & Train & Valid & Test &  Distill & Train & Valid & Test\\
    \hline
    \# of pairs & 300M & 393K & 12.8K & 12.8K & 40M & 1.1M & 12.8K & 12.8K  \\
    \hline
    AvgLen product/passage & 107.6 & 84.3 & 83.5 & 82.1 & 55.5 & 56.0 & 53.6 & 53.8  \\
    \hline
    AvgLen term/query & 1.49 & 1.32& 1.32& 1.31 & 6.37 & 6.03 & 6.00 & 6.03  \\
    \hline \hline
    \end{tabular}
    \vspace{-1em}
  \caption{Datasets statistics.}
  \label{tb:datasets}
\end{table*}

\subsection{Baseline Approaches}
\label{sec:baseline}
There exist many knowledge distillation (see Sec. \ref{sec:related-work} for more details) works, but none of them has been optimized for pairwise input. We choose to compare our DiPair approach with the fastest BERT-based student model \cite{DBLP:journals/corr/abs-1908-08962} we are aware of, and our model is at least \textbf{8x} faster (see Table \ref{tb:main-q2p-bert}). We also compare our proposed approach with several other strong baselines:
\vspace{-0.7em}
\begin{itemize}
\itemsep-0.3em
    \item \berttiny: the fastest version of BERT released in \cite{DBLP:journals/corr/abs-1908-08962}. This model has $2$ layers with $128$D word embeddings and $2$-head transformer. It is claimed to be $52$x faster than BERT-base (on TPU), and to the best of our knowledge, this is faster than any other BERT-based student models in the literature.
    \item \decos: BERT-based Dual-Encoder model. Cosine between left/right CLS embeddings is used as the similarity score. 
    \item \deffnn: BERT-based Dual-Encoder model. FFNN (Feedforward Neural Networks) is used to aggregate the left/right CLS embeddings into a similarity score. Unless otherwise stated, we fix the FFNN to be 2-Layer with dimensions x128x128. The input to the FFNN has dimension $768 + 768 =1536$. 
    \item \denmtsf: our proposed model, BERT-based Dual-Encoder, with a transformer-based head.  $N$ and $M$ refer to the output sequence lengths (see Figure \ref{fig:di-pair-model}). In all experiments, we fix our head to be 2-Layer, 1-Head, 1024D intermediate size. The value of hidden\_size (i.e.,  the dimension of the input token embeddings) is decided by the output of the projection layer.
    \item \denmffnn: this is similar to \denmtsf; the only difference is that the transformer-based head is replaced with an FFNN. The input to the FFNN has dimension $(N + M)$ * hidden\_size (N, M defined in Figure \ref{fig:di-pair-model}
    ).  We use 2-Layer FFNN with dimensions x128x128 unless otherwise stated. 
\end{itemize}
In all the aforementioned models (except \berttiny), the dual-encoder is initialized from the first $K$ layers of the pre-trained BERT model as well as the token embedding matrix.  Unless otherwise stated, we fix K=1 for \ptdata~and K=4 for \qpdata.
The Left encoder and the right encoder will share parameters. For models with a projection layer, we use $D$ to represent the dimension of the projected result.

\subsection{Model Distillation}
\label{sec:model_distillation}
\paragraph{Teacher Models}
For \qpdata, we use Google's public  12-layer BERT-base pre-trained model, and fine-tune it with the 1.1M labeled query to passage pairs.

On the other hand, for \ptdata~ data, we pre-train a 12-layer BERT-based model with a customized vocabulary of size 80K, using user interaction data. We use the default parameters released in the public BERT code.\footnote{Available in \url{https://github.com/google-research/bert}.}  We then fine-tune the pre-trained model using the 393K product to term pairs.

For both teachers, we use the following cross-entropy loss,
\vspace{-0.5em}
\begin{equation}
\label{eq:loss}
     -\sum_i \left( y_i \log p_i + (1 - y_i) \log (1 - p_i) \right)
     \vspace{-.7em}
\end{equation}
where $y_i$ is the label and $p_i$ is computed via applying a sigmoid function on the teacher's logits $z_i$. This loss function works for both regression problems and binary classification problems.
\paragraph{Distillation}
Inspired by \citet{DBLP:journals/corr/HintonVD15}, we use $\text{sigmoid}(z_i/T)$ to create soft labels to annotate the distillation sets, where $z_i$ is teacher's logits and $T$ is known as the temperature. In our experiment, we fix $T=1$. We then apply the cross entropy loss as detailed in Equation (\ref{eq:loss}).

\subsection{Experimental Setup}
Our code is implemented with TensorFlow \footnote{\url{https://www.tensorflow.org/}} and we use TPUv3 in all of our experiments. We use AdamW optimizer following the public BERT code. The warmup step is fixed to be 50k. Other parameters of the optimizer are identical to the default values set in the public BERT code (  weight\_decay\_rate=0.01,
      $\beta_1=0.9$,
      $\beta_2=0.999$,
      $\epsilon=1e^{-6}$).

We tune some other key hyper-parameters using the validation sets. We try multiple (learning rate, batch size) combinations and choose the best ones. In the two-stage training, the models are less sensitive to learning rates in the first stage, and we set the learning rate as 5e-5;  we then train the models until they converge. In the second stage of training, the learning rate is set to be 5e-5 in \denmtsf, \decos, \denmffnn; we use batch size 512 and 4x4 TPU topology. For \berttiny, we use batch size 128, learning rate 2e-6, and 2x2 TPU topology. 
 All other hyperparameters related to model architecture are specified in Sec. \ref{sec:baseline}.


\begin{table*}[!ht]
\vspace{-1em}
  \begin{subtable}[h]{1\textwidth}
    \centering
    \small
    \begin{tabular}{c|c|c|c|c|c}
        \hline \hline
        Model Settings & Pearson (valid) & Delta (valid) & Pearson (test) & Delta (test) & Speedup \\
        \hline \hline
        Teacher (BERT-base) & 0.757 & -0\% & 0.757 & -0\% & 1x \\
        \hline
        \hline
        \deffnn & 0.682 & -9.9\% & 0.677 & -11.6\% & 3129x \\
        \hline 
        \decos & 0.678 & -10.4\% & 0.669 & -11.6\% & \textbf{3990x} \\
        \hline
        \denmffnn& 0.696 & -8.1\% & 0.697 & -7.9\% & 2128x \\
        \hline \hline
        \denmtsf & \textbf{0.732}* & \textbf{-3.3\%}* & \textbf{0.731}* & \textbf{-3.4\%}* & 362x \\
        \hline\hline
    \end{tabular}
    \caption{Comparison with Dual-Encoder based model.}
   \end{subtable}
   \begin{subtable}[h]{1\textwidth}
    \centering
    \small
    \begin{tabular}{c|c|c|c|c|c}
        \hline \hline
        Model Settings & Pearson (valid) & Delta (valid) & Pearson (test) & Delta (test) & Speedup \\
        \hline \hline
        ~~~~~\berttiny~~~~~~~~ & 0.644  & -12.0\% & 0.640 & -11.3\% & 53x \\
        \hline \hline
        \denmtsf & \textbf{0.732}* & \textbf{-3.3\%}* & \textbf{0.731}* & \textbf{-3.4\%}* & \textbf{362x} \\
        \hline\hline
    \end{tabular}
    \caption{Comparison with BERT-based student model.}
  \end{subtable} 
  \vspace{-1em}
    \caption{Main results for \ptdata~data. Entries marked with * are significant (p-value $<$ 0.05, w.r.t. the closest baseline, following \cite{BBK12}). For \denmtsf~and \denmffnn, we set N=4, M=12 and projected dimension D=128. Both teacher model and \berttiny~ take input with length 128.
    The teacher model is a customized BERT model, with a vocabulary of size 80K.
   \berttiny~has a different vocab, this explains why it has the worst performance. We report the running time of the heads (measured on CPU), as \#pairs $\gg$ \#products $+$ \#terms.}
    \label{tb:main-p2t}
\end{table*}

\begin{table*}[!ht]
  \begin{subtable}[h]{1\textwidth}
    \centering
    \small
    \begin{tabular}{c|c|c|c|c|c}
        \hline\hline
        Model Settings & AUC\_ROC (valid) & Delta (valid) & AUC\_ROC (test) & Delta (test) & Speedup \\
        \hline\hline
        Teacher (BERT-base) & 0.955 & -0\% & 0.957 & -0\% & 1x \\
        \hline
        \hline
 
        \deffnn & 0.895 &  -6.3\% & 0.896 & -6.4\% & 3863x \\
        \hline
        \decos & 0.871 & -8.8\% & 0.878 &  -8.3\% & \textbf{5109x} \\
        \hline
        \denmffnn & 0.900 & -5.8\% & 0.904 & -5.5\% & 2437x \\
        \hline\hline
        \denmtsf & \textbf{0.930}* & \textbf{-2.6\%}* & \textbf{0.932}*  & \textbf{-2.6\%}* & 355x \\
        \hline\hline
    \end{tabular}
    \caption{Comparison with Dual-Encoder based models.}
    \label{tb:main-q2p-de}
  \end{subtable}
  \hfill
  \begin{subtable}[h]{1\textwidth}
    \centering
    \small
    \begin{tabular}{c|c|c|c|c|c}
    \hline\hline
        Model Settings & AUC\_ROC (valid) & Delta (valid) & AUC\_ROC (test) & Delta (test) & Speedup \\
        \hline\hline
        ~~~~~\berttiny~~~~~~~ & \textbf{0.933}* & \textbf{-2.3\%}*  & \textbf{0.936}* & \textbf{-2.2\%}* & 44x \\
        \hline
        \denmtsf  & 0.930 & -2.6\%  & 0.932 & -2.6\% & \textbf{355x} \\
        \hline\hline
    \end{tabular}
    \caption{Comparison with BERT-based student model.}
    \label{tb:main-q2p-bert}
  \end{subtable}
      \caption{Main results for \qpdata ~data. Entries marked with * are significant (p-value $<$ 0.05, w.r.t. the closest baseline, following the approach detailed in \cite{BBK12}). For \denmtsf~and \denmffnn, N=4, M=8, D=256. The input to the teacher model and \berttiny~ has  length 128. Query encoder and passage encoder take input with lengths 32 and 128, respectively.}
    \label{tb:main-q2p}
\end{table*}

\begin{table*}[!ht]
    \vspace{-.5em}
    \centering
    \small
    \begin{tabular}{c|c|c|c|c|c|c|c}
        \hline\hline
         \# & Model Type & Head Settings & $N$ & $M$ & \#Params in Head & AUC\_ROC & Speedup\\
        \hline
        0 & Teacher & - & - & - & -& 0.955 & 1x \\
        \hline\hline
         1 & \denmtsf & 2-Layer & 4 & 8 & 1.7M & 0.930 & 355x \\
          \hline
        2 &  \denmtsf & 2-Layer& 8 & 16 & 1.7M & 0.942 & 98x \\
          \hline\hline
        3 &  \denmffnn & x$2^7$x$2^7$ & 4 & 8 & 0.4M & 0.900 & 2437x \\
         \hline
        4 & \denmffnn & x$2^{10}$x$2^{10}$ & 4 & 8 & 4.2M & 0.909 & 616x \\
        \hline
        5 & \denmffnn & x$2^{10}$x$2^{10}$ & 8 &16 & 7.3M &  0.908 & 268x \\
         \hline\hline
        6 &  \deffnn & x$2^7$x$2^7$ &- &- & 0.2M & 0.895 & 3863x \\
          \hline
        7 &  \deffnn & x$2^{10}$x$2^{10}$ &- &- & 2.6M  & 0.912 & 754x \\
         \hline
        8 & \deffnn & x$2^{10}$x$2^{10}$x$2^{10}$x$2^{10}$ &- &- & 4.7M & 0.909 & 420x \\
        \hline\hline
    \end{tabular}
    \caption{Varying the head settings in \deffnn, \denmffnn~and \denmtsf. \#Params refers to the number of trainable parameters in the head. We set D=256 in \denmtsf~and \denmffnn. \#Params is independent of $N$ and $M$ in \denmtsf, but not in \denmffnn.}
    \label{tb:head}
\end{table*}

\subsection{Main Results}
\label{sec:main_results}





Table \ref{tb:main-p2t} and Table \ref{tb:main-q2p} present the experimental results on \ptdata~and \qpdata~datasets, respectively.
Among all the student models with dual-encoder architecture,  \denmtsf~consistently achieves the best performance. For the \qpdata~dataset, \denmtsf~ achieves similar AUC\_ROC to \berttiny; however, it achieves a \textbf{8x} speedup.

Among all the student models, \decos~is the fastest one as it only requires dot product during inference. However, it has the worst performance, indicating that using Cosine function alone does not allow enough interaction between the input sequences embeddings.


\subsection{Effectiveness of Transformer Head}
\label{sec:head}
To verify the importance of using a transformer-based head, we vary \#params in the heads of \denmtsf, \deffnn~and \denmffnn. Table \ref{tb:head} presents the experimental results.

Comparing rows 1 and 2 in Table \ref{tb:head}, the model quality of \denmtsf~ can be improved by increasing the head input sequences lengths ($N$ and $M$), although at the cost of longer inference time. On the other hand, rows 3-5 show that increasing \#Params in FFNN head (e.g., using larger dimensions, more layers) does not lead to significant quality improvement for \denmffnn;  even when the \#Params of the FFNN head is 4x more than the transformer head, the model quality of \denmtsf~ is still considerably superior to that of \denmffnn(cf. rows 2 and 5). A similar conclusion can be made for \deffnn~(rows 6-8).

Another interesting observation is that even with more input information and more parameters, \denmffnn~ does not generate higher AUC\_ROC than \deffnn. This might suggest that FFNN is not powerful enough to aggregate the input information effectively.

Overall, Table \ref{tb:head} illustrates the importance of using a transformer head if we want to achieve high model quality: Unlike FFNN-based heads, where we could not further improve the model via increasing \#Params, a transformer-based head has more headroom to reduce the distillation gap further, and the desired quality-speed trade-off can be easily achieved by adjusting the values of $N$ and $M$.

\subsection{Effect of Two-Stage Training}
\label{exp:two-stage}
\begin{figure}[ht]
     \centering
     \small
     \vspace{-1em}
     \hspace{-10em}
     \begin{subfigure}{0.6\columnwidth}
         \centering
         \includegraphics[width=0.9\columnwidth]{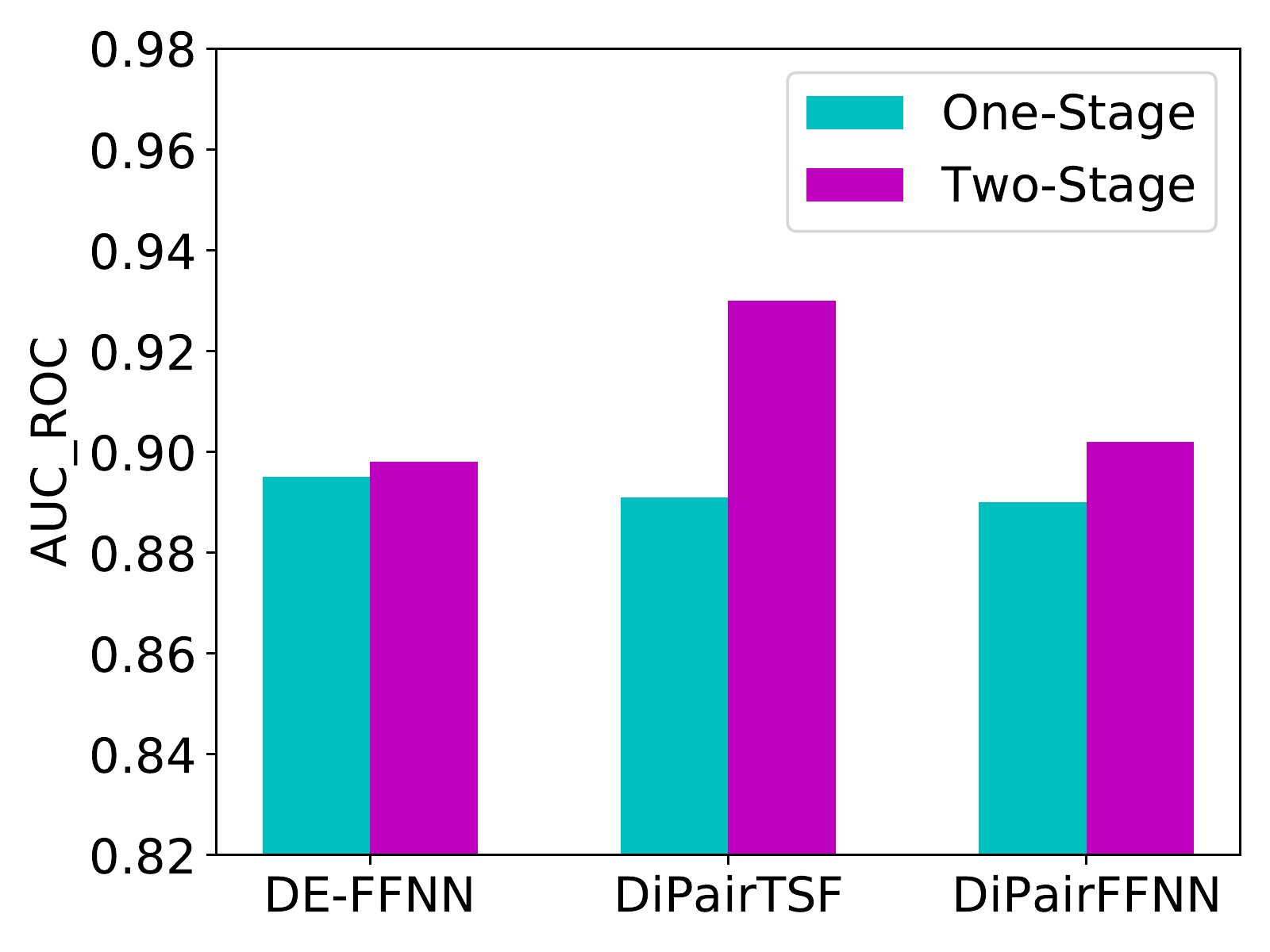}
         \caption{}
         \label{fig:two-stage}
     \end{subfigure}
     \hspace{-1.4em}
     \begin{subfigure}{0.6\columnwidth}
         \centering
         \small
         \includegraphics[width=0.9\columnwidth]{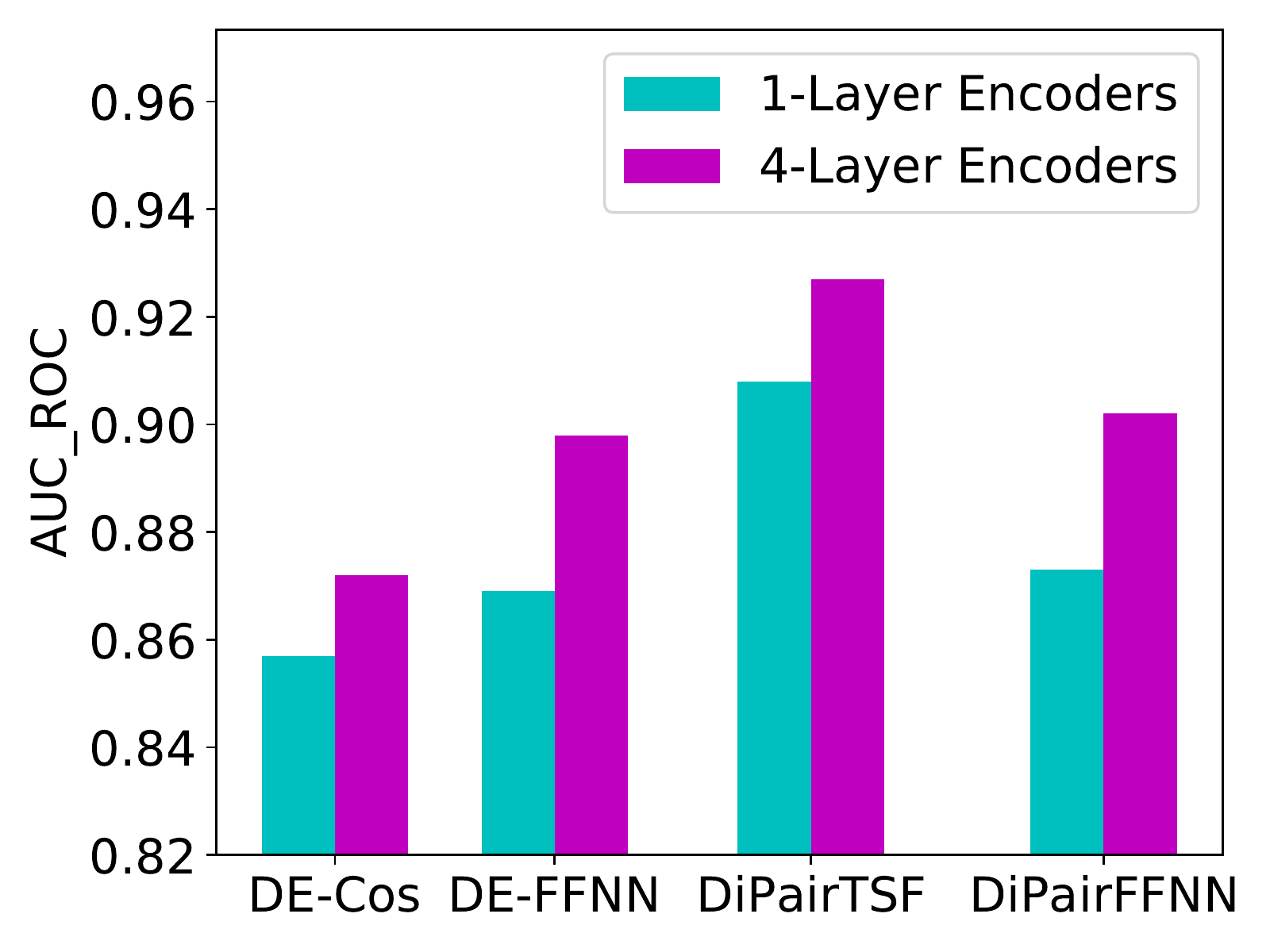}
         \caption{}
         \label{fig:encoder-layers}
     \end{subfigure}
     \hspace{-10em}
     \vspace{-1em}
       \caption{(a) One-stage training v.s. two-stage training. Projection dimension D=256. (b) Different \# of layers in encoders. D=256}
        \label{fig:ablation}
    \vspace{-1.5em}
\end{figure}

Figure \ref{fig:two-stage} shows that two-stage training, which is discussed in Section \ref{sec:twostage} has positive effects on all the methods we test. When the head is transformer-based, the two-stage training plays an important role: the AUC\_ROC improves from $0.891$ to $0.930$. 

On the other hand, the gain introduced by using two-stage training is less significant in other approaches such as \deffnn~ and \denmffnn. This might be because FFNN is generally easier to train than transformer-based models, and thus initialization choices play a lesser role.

\subsection{Model Ablation Studies}
\label{sec:model-ablation}




\paragraph{Varying the Encoder Layers}
Figure \ref{fig:encoder-layers} shows that we can improve the model performance by increasing the number of layers in the encoders. Since the heads remain the same, and the number of pairs is often far greater than the number of the unique items needed to be encoded, the total inference time will not increase accordingly.

\begin{figure}[t]
    \centering
    \vspace{-0.1em}
    \includegraphics[width=0.48\columnwidth]{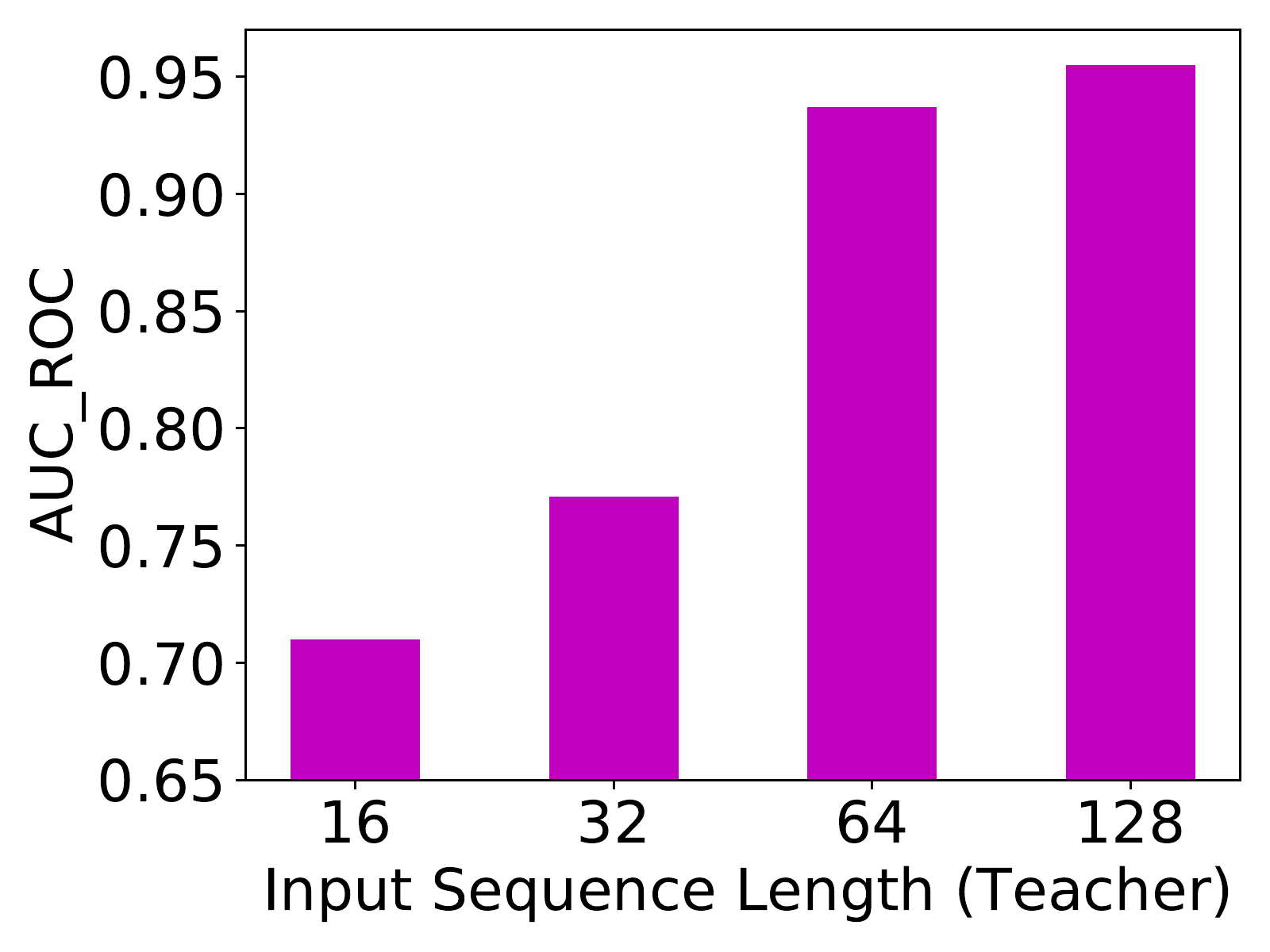}
    \vspace{-1em}
    \caption{The effect of input sequence length.}
    \label{fig:reduce-seq-len}
    \vspace{-.0em}
\end{figure}

\paragraph{Reducing Input Sequence Length}
Figure \ref{fig:reduce-seq-len} shows that if we reduce the input sequence length in BERT, the quality of the model drops quickly as there is not enough information available for the model to make the correct decision.

\paragraph{Dimension of the Projection Layer}
We vary the projection dimension D. Table \ref{tb:dim-projection} shows that AUC\_ROC drops quickly when we aggressively reduce D from 256 to 16. This is expected as less information can be preserved with a smaller projection dimension. On the other hand, removing projection layer completely leads to almost no improvement over the 256D version. This indicates that adding projection layer is a useful strategy to save both storage and running time, without hurting the model quality.
\begin{table}[!ht]
    \centering
    \small
    \vspace{-0.5em}
    \begin{tabular}{|c|c|}
        \hline
        Output Dim of Projection  & AUC\_ROC \\
        \hline
         256D & 0.930 \\
        \hline
         128D & 0.904 \\
        \hline
         16D & 0.831 \\
        \hline
        No projection, 768D & 0.930 \\
        \hline
    \end{tabular}
    \vspace{-.7em}
    \caption{The effect of projection layer for \denmtsf.}
    \label{tb:dim-projection}
    \vspace{-1.5em}
\end{table}

\paragraph{First N + M Tokens v.s. Last N + M}
Since our \denmtsf~model is end to end trained, the model should learn to push the information of the full input sequence to \emph{arbitrarily} selected (N + M) token embeddings. To verify this intuition, we select the last (N + M) token embeddings from the dual-encoder output and compare it with the one using the first (N + M). As expected, when we fix N=4, M=8,  replacing the first tokens with the last tokens only changes  AUC\_ROC from 0.930 to 0.925, which is almost neglectable.

\paragraph{Effect of Output Sequence Lengths}
\label{exp:out-seq-len}
Table \ref{tb:output-seq-len} illustrates that for a transformer-based head, the model quality drops when we reduce the output sequence lengths ($8 \rightarrow 2$, $16\rightarrow 2$). Here we fix D=256.

Another observation is that (N=11, M=1) is worse than any other configurations with the same value of (N+M).
This might because in this \qpdata~data, queries are usually shorter than the passages, and we might need more token embeddings to store the information of a passage; therefore, M should greater than 1.

\begin{table}[!h]
    \centering
    \small
    \begin{tabular}{|c|c|c|c|c|}
        \hline
         N & M & L & AUC\_ROC \\
        \hline
          8 & 16 & 2 & 0.942 \\
         \hline
          8 & 4 & 2 & 0.934 \\
          \hline
          4 & 8 & 4 & 0.936 \\
         \hline
         4 & 8 & 2 & 0.930 \\
        \hline
         2 & 2 & 2 & 0.909 \\
         \hline
         1 & 11 & 2 & 0.922 \\
         \hline
         11 & 1 & 2 & 0.916 \\
        \hline
    \end{tabular}
    \caption{The effect of output sequence lengths in \denmtsf. L is the \#layers in the transformer head.}
    \label{tb:output-seq-len}
    \vspace{-0.8em}
\end{table}




\section{Open Questions}
\label{sec:open-questions}
\vspace{-0.4em}

DiPair has been discussed in the context of knowledge distillation in this work, but it can be trivially extended to more scenarios, as we can train it directly. The proposed framework raises several research questions.
\paragraph{Learning Dynamics of Our Model}
Recall that, in our framework, each encoder outputs its first few token embeddings as the input to the head, and we end to end to train the model to force the encoder to push the information of the input text into those outputted embeddings. However, it is unclear to us what those outputted embeddings actually learn. It would be interesting to understand the learning dynamics of our model.

\paragraph{Models for Online Serving}
In some applications, we are interested in serving the model online. Our proposed framework uses transformer-based encoders and requires to pre-compute the embeddings. As a result, it is difficult to serve our model online. It can be extremely useful to extend our framework for online use cases. Here we give a more concrete example: To score the query to document relevance online, we can usually pre-compute the embeddings of documents and index them, so using an expensive document encoder is not an issue; however, the query encoder and the head must be run online.

\paragraph{Extension to Non-Textual Features}
Another interesting situation to consider is when one side (or both sides) of the input pair is non-textual. For example, we may care about scoring a pair of (image, document), or a pair of (audio, document). Such applications require us to modify our proposed architecture to better fit non-textual features.

\section{Conclusion and Future Work}
\vspace{-0.4em}
In this work, we reveal the importance of customizing models for problems with pairwise/n-ary input and propose a new framework, DiPair, as an effective solution. This framework is flexible, and we can easily achieve more than 350x speedup over a BERT-based teacher model with no significant quality drop.

\section*{Acknowledgments}

We would like to thank Krishna Srinivasan for his feedback and suggestions.
We would also like to thank Anand Murugappan, Corinna Cortes and Greg Friedman for their support.
\clearpage
\bibliographystyle{acl_natbib}
\bibliography{dipair2020}

\begin{thebibliography}{39}
\expandafter\ifx\csname natexlab\endcsname\relax\def\natexlab#1{#1}\fi

\bibitem[{Berg{-}Kirkpatrick et~al.(2012)Berg{-}Kirkpatrick, Burkett, and
  Klein}]{BBK12}
T.~Berg{-}Kirkpatrick, D.~Burkett, and D.~Klein. 2012.
\newblock \href {https://www.aclweb.org/anthology/D12-1091/} {An empirical
  investigation of statistical significance in {NLP}}.
\newblock In \emph{Proceedings of the 2012 Joint Conference on Empirical
  Methods in Natural Language Processing and Computational Natural Language
  Learning, EMNLP-CoNLL 2012, July 12-14, 2012, Jeju Island, Korea}, pages
  995--1005. {ACL}.

\bibitem[{Bowman et~al.(2015)Bowman, Angeli, Potts, and
  Manning}]{bowman-etal-2015-large}
S.~R. Bowman, G.~Angeli, C.~Potts, and C.~D. Manning. 2015.
\newblock \href {https://doi.org/10.18653/v1/D15-1075} {A large annotated
  corpus for learning natural language inference}.
\newblock In \emph{Proceedings of the 2015 Conference on Empirical Methods in
  Natural Language Processing}, pages 632--642, Lisbon, Portugal. Association
  for Computational Linguistics.

\bibitem[{Cer et~al.(2018)Cer, Yang, Kong, Hua, Limtiaco, St.~John, Constant,
  Guajardo-Cespedes, Yuan, Tar, Strope, and Kurzweil}]{cer-etal-2018-universal}
D.~Cer, Y.~Yang, S.~Kong, N.~Hua, N.~Limtiaco, R.~St.~John, N.~Constant,
  M.~Guajardo-Cespedes, S.~Yuan, C.~Tar, B.~Strope, and R.~Kurzweil. 2018.
\newblock \href {https://doi.org/10.18653/v1/D18-2029} {Universal sentence
  encoder for {E}nglish}.
\newblock In \emph{Proceedings of the 2018 Conference on Empirical Methods in
  Natural Language Processing: System Demonstrations}, pages 169--174,
  Brussels, Belgium. Association for Computational Linguistics.

\bibitem[{Chidambaram et~al.(2019)Chidambaram, Yang, Cer, Yuan, Sung, Strope,
  and Kurzweil}]{chidambaram-etal-2019-learning}
M.~Chidambaram, Y.~Yang, D.~Cer, S.~Yuan, Y.~Sung, B.~Strope, and R.~Kurzweil.
  2019.
\newblock \href {https://doi.org/10.18653/v1/W19-4330} {Learning cross-lingual
  sentence representations via a multi-task dual-encoder model}.
\newblock In \emph{Proceedings of the 4th Workshop on Representation Learning
  for NLP (RepL4NLP-2019)}, pages 250--259, Florence, Italy. Association for
  Computational Linguistics.

\bibitem[{Clark et~al.(2020)Clark, Luong, Le, and Manning}]{electra20}
K.~Clark, M.~Luong, Q.~V. Le, and C.~D. Manning. 2020.
\newblock \href {https://openreview.net/forum?id=r1xMH1BtvB} {{ELECTRA:}
  pre-training text encoders as discriminators rather than generators}.
\newblock In \emph{8th International Conference on Learning Representations,
  {ICLR} 2020, Addis Ababa, Ethiopia, April 26-30, 2020}. OpenReview.net.

\bibitem[{Das et~al.(2016)Das, Yenala, Chinnakotla, and
  Shrivastava}]{das-etal-2016-together}
A.~Das, H.~Yenala, M.~Chinnakotla, and M.~Shrivastava. 2016.
\newblock \href {https://doi.org/10.18653/v1/P16-1036} {Together we stand:
  {S}iamese networks for similar question retrieval}.
\newblock In \emph{Proceedings of the 54th Annual Meeting of the Association
  for Computational Linguistics (Volume 1: Long Papers)}, pages 378--387,
  Berlin, Germany. Association for Computational Linguistics.

\bibitem[{Devlin et~al.(2018)Devlin, Chang, Lee, and
  Toutanova}]{devlin2018:bert}
J.~Devlin, M.~Chang, K.~Lee, and K.~Toutanova. 2018.
\newblock \href {http://arxiv.org/abs/1810.04805} {{BERT:} pre-training of deep
  bidirectional transformers for language understanding}.
\newblock \emph{CoRR}, abs/1810.04805.

\bibitem[{Frankle and Carbin(2019)}]{FC19}
J.~Frankle and M.~Carbin. 2019.
\newblock \href {https://openreview.net/forum?id=rJl-b3RcF7} {The lottery
  ticket hypothesis: Finding sparse, trainable neural networks}.
\newblock In \emph{7th International Conference on Learning Representations,
  {ICLR} 2019, New Orleans, LA, USA, May 6-9, 2019}. OpenReview.net.

\bibitem[{Guo et~al.(2016)Guo, Fan, Ai, and
  Croft}]{Guo:2016:DRM:2983323.2983769}
J.~Guo, Y.~Fan, Q.~Ai, and W.~B. Croft. 2016.
\newblock A deep relevance matching model for ad-hoc retrieval.
\newblock In \emph{CIKM '16}.

\bibitem[{Guo et~al.(2019)Guo, Fan, Pang, Yang, Ai, Zamani, Wu, Croft, and
  Cheng}]{DBLP:journals/corr/abs-1903-06902}
J.~Guo, Y.~Fan, L.~Pang, L.~Yang, Q.~Ai, H.~Zamani, C.~Wu, W.~B. Croft, and
  X.~Cheng. 2019.
\newblock \href {http://arxiv.org/abs/1903.06902} {A deep look into neural
  ranking models for information retrieval}.

\bibitem[{Han et~al.(2016)Han, Mao, and Dally}]{HMD16}
S.~Han, H.~Mao, and W.~J. Dally. 2016.
\newblock \href {http://arxiv.org/abs/1510.00149} {Deep compression:
  Compressing deep neural network with pruning, trained quantization and
  huffman coding}.
\newblock In \emph{4th International Conference on Learning Representations,
  {ICLR} 2016, San Juan, Puerto Rico, May 2-4, 2016, Conference Track
  Proceedings}.

\bibitem[{Henderson et~al.(2017)Henderson, Al{-}Rfou, Strope, Sung,
  Luk{\'{a}}cs, Guo, Kumar, Miklos, and
  Kurzweil}]{DBLP:journals/corr/HendersonASSLGK17}
M.~L. Henderson, R.~Al{-}Rfou, B.~Strope, Y.~Sung, L.~Luk{\'{a}}cs, R.~Guo,
  S.~Kumar, B.~Miklos, and R.~Kurzweil. 2017.
\newblock \href {http://arxiv.org/abs/1705.00652} {Efficient natural language
  response suggestion for smart reply}.
\newblock \emph{CoRR}, abs/1705.00652.

\bibitem[{Hinton et~al.(2015)Hinton, Vinyals, and
  Dean}]{DBLP:journals/corr/HintonVD15}
G.~E. Hinton, O.~Vinyals, and J.~Dean. 2015.
\newblock \href {http://arxiv.org/abs/1503.02531} {Distilling the knowledge in
  a neural network}.
\newblock \emph{CoRR}, abs/1503.02531.

\bibitem[{Howard et~al.(2017)Howard, Zhu, Chen, Kalenichenko, Wang, T.Weyand,
  Andreetto, and Adam}]{HZK+17}
A.~G. Howard, M.~Zhu, B.~Chen, D.~Kalenichenko, W.~Wang, T.Weyand,
  M.~Andreetto, and H.~Adam. 2017.
\newblock \href {http://arxiv.org/abs/1704.04861} {Mobilenets: Efficient
  convolutional neural networks for mobile vision applications}.
\newblock \emph{CoRR}, abs/1704.04861.

\bibitem[{Hu et~al.(2014)Hu, Lu, Li, and Chen}]{DBLP:conf/nips/HuLLC14}
B.~Hu, Z.~Lu, H.~Li, and Q.~Chen. 2014.
\newblock Convolutional neural network architectures for matching natural
  language sentences.
\newblock In \emph{NIPS '14}.

\bibitem[{Huang et~al.(2013)Huang, He, Gao, Deng, Acero, and
  Heck}]{DBLP:conf/cikm/HuangHGDAH13}
P.~Huang, X.~He, J.~Gao, L.~Deng, A.~Acero, and L.~P. Heck. 2013.
\newblock Learning deep structured semantic models for web search using
  clickthrough data.
\newblock In \emph{CIKM '13}.

\bibitem[{Humeau et~al.(2020)Humeau, Shuster, Lachaux, and Weston}]{HSL+20}
S.~Humeau, K.~Shuster, M.~Lachaux, and J.~Weston. 2020.
\newblock \href {https://openreview.net/forum?id=SkxgnnNFvH} {Poly-encoders:
  Architectures and pre-training strategies for fast and accurate
  multi-sentence scoring}.
\newblock In \emph{8th International Conference on Learning Representations,
  {ICLR} 2020, Addis Ababa, Ethiopia, April 26-30, 2020}. OpenReview.net.

\bibitem[{Iandola et~al.(2016)Iandola, Moskewicz, Ashraf, Han, Dally, and
  Keutzer}]{IMA+16}
F.~N. Iandola, M.~W. Moskewicz, K.~Ashraf, S.~Han, W.J. Dally, and K.~Keutzer.
  2016.
\newblock \href {http://arxiv.org/abs/1602.07360} {Squeezenet: Alexnet-level
  accuracy with 50x fewer parameters and {\textless}1mb model size}.
\newblock \emph{CoRR}, abs/1602.07360.

\bibitem[{Jiao et~al.(2019)Jiao, Yin, Shang, Jiang, Chen, Li, Wang, and
  Liu}]{DBLP:journals/corr/abs-1909-10351}
X.~Jiao, Y.~Yin, L.~Shang, X.~Jiang, X.~Chen, L.~Li, F.~Wang, and Q.~Liu. 2019.
\newblock \href {http://arxiv.org/abs/1909.10351} {Tinybert: Distilling {BERT}
  for natural language understanding}.
\newblock \emph{CoRR}, abs/1909.10351.

\bibitem[{Johnson et~al.()Johnson, Douze, and J{\'e}gou}]{JDH17}
J.~Johnson, M.~Douze, and H.~J{\'e}gou.
\newblock Billion-scale similarity search with gpus.

\bibitem[{Lan et~al.(2019)Lan, Chen, Goodman, Gimpel, Sharma, and
  Soricut}]{albert19}
Z.~Lan, M.~Chen, S.~Goodman, K.~Gimpel, P.~Sharma, and R.~Soricut. 2019.
\newblock \href {http://arxiv.org/abs/1909.11942} {{ALBERT:} {A} lite {BERT}
  for self-supervised learning of language representations}.
\newblock \emph{CoRR}, abs/1909.11942.

\bibitem[{Li and Xu(2014)}]{10.5555/2683840}
H.~Li and J.~Xu. 2014.
\newblock \emph{Semantic Matching in Search}.
\newblock Now Publishers Inc., Hanover, MA, USA.

\bibitem[{Liu et~al.(2019)Liu, Ott, Goyal, Du, Joshi, Chen, O.Levy, Lewis,
  Zettlemoyer, and Stoyanov}]{roberta19}
Y.~Liu, M.~Ott, N.~Goyal, J.~Du, M.~Joshi, D.~Chen, O.Levy, M.~Lewis,
  L.~Zettlemoyer, and V.~Stoyanov. 2019.
\newblock \href {http://arxiv.org/abs/1907.11692} {Roberta: {A} robustly
  optimized {BERT} pretraining approach}.
\newblock \emph{CoRR}, abs/1907.11692.

\bibitem[{MacAvaney et~al.(2020)MacAvaney, Nardini, Perego, Tonellotto,
  Goharian, and Frieder}]{macavaney2020efficient}
S.~MacAvaney, F.~Maria Nardini, R.~Perego, N.~Tonellotto, N.~Goharian, and
  O.~Frieder. 2020.
\newblock \href {http://arxiv.org/abs/2004.14255} {Efficient document
  re-ranking for transformers by precomputing term representations}.

\bibitem[{Mitra et~al.(2017)Mitra, Diaz, and
  Craswell}]{Mitra:2017:LMU:3038912.3052579}
B.~Mitra, F.~Diaz, and N.~Craswell. 2017.
\newblock Learning to match using local and distributed representations of text
  for web search.
\newblock In \emph{WWW '17}.

\bibitem[{Pang et~al.(2016)Pang, Lan, Guo, Xu, Wan, and
  Cheng}]{DBLP:conf/aaai/PangLGXWC16}
L.~Pang, Y.~Lan, J.~Guo, J.~Xu, S.~Wan, and X.~Cheng. 2016.
\newblock Text matching as image recognition.
\newblock In \emph{AAAI '16}.

\bibitem[{Rao et~al.(2019)Rao, Liu, Tay, Yang, Shi, and
  Lin}]{rao-etal-2019-bridging}
J.~Rao, L.~Liu, Y.~Tay, W.~Yang, P.~Shi, and J.~Lin. 2019.
\newblock \href {https://doi.org/10.18653/v1/D19-1540} {Bridging the gap
  between relevance matching and semantic matching for short text similarity
  modeling}.
\newblock In \emph{{EMNLP-IJCNLP} 2019}, pages 5370--5381, Hong Kong, China.
  Association for Computational Linguistics.

\bibitem[{Reimers and Gurevych(2019)}]{reimers-gurevych-2019-sentence}
N.~Reimers and I.~Gurevych. 2019.
\newblock \href {https://doi.org/10.18653/v1/D19-1410} {Sentence-{BERT}:
  Sentence embeddings using {S}iamese {BERT}-networks}.
\newblock In \emph{Proceedings of the 2019 Conference on Empirical Methods in
  Natural Language Processing and the 9th International Joint Conference on
  Natural Language Processing (EMNLP-IJCNLP)}, pages 3982--3992, Hong Kong,
  China. Association for Computational Linguistics.

\bibitem[{Renda et~al.(2020)Renda, Frankle, and M.Carbin}]{RFC20}
A.~Renda, J.~Frankle, and M.Carbin. 2020.
\newblock \href {https://openreview.net/forum?id=S1gSj0NKvB} {Comparing
  rewinding and fine-tuning in neural network pruning}.
\newblock In \emph{8th International Conference on Learning Representations,
  {ICLR} 2020, Addis Ababa, Ethiopia, April 26-30, 2020}. OpenReview.net.

\bibitem[{Sanh et~al.(2019)Sanh, Debut, Chaumond, and
  Wolf}]{DBLP:journals/corr/abs-1910-01108}
V.~Sanh, L.~Debut, J.~Chaumond, and T.~Wolf. 2019.
\newblock \href {http://arxiv.org/abs/1910.01108} {Distilbert, a distilled
  version of {BERT:} smaller, faster, cheaper and lighter}.
\newblock \emph{CoRR}, abs/1910.01108.

\bibitem[{Sun et~al.(2019)Sun, Cheng, Gan, and Liu}]{DBLP:conf/emnlp/SunCGL19}
S.~Sun, Y.~Cheng, Z.~Gan, and J.~Liu. 2019.
\newblock \href {https://doi.org/10.18653/v1/D19-1441} {Patient knowledge
  distillation for {BERT} model compression}.
\newblock In \emph{{EMNLP-IJCNLP} 2019, Hong Kong, China, November 3-7, 2019},
  pages 4322--4331. Association for Computational Linguistics.

\bibitem[{Tang et~al.(2019)Tang, Lu, Liu, Mou, Vechtomova, and
  Lin}]{DBLP:journals/corr/abs-1903-12136}
R.~Tang, Y.~Lu, L.~Liu, L.~Mou, O.~Vechtomova, and J.~Lin. 2019.
\newblock \href {http://arxiv.org/abs/1903.12136} {Distilling task-specific
  knowledge from {BERT} into simple neural networks}.
\newblock \emph{CoRR}, abs/1903.12136.

\bibitem[{Turc et~al.(2019)Turc, Chang, Lee, and
  Toutanova}]{DBLP:journals/corr/abs-1908-08962}
I.~Turc, M.~Chang, K.~Lee, and K.~Toutanova. 2019.
\newblock \href {http://arxiv.org/abs/1908.08962} {Well-read students learn
  better: The impact of student initialization on knowledge distillation}.
\newblock \emph{CoRR}, abs/1908.08962.

\bibitem[{Vaswani et~al.(2017)Vaswani, Shazeer, Parmar, Uszkoreit, Jones,
  Gomez, Kaiser, and Polosukhin}]{NIPS2017_Transformers}
A.~Vaswani, N.~Shazeer, N.~Parmar, J.~Uszkoreit, L.~Jones, A.~N. Gomez, \L~.
  Kaiser, and I.~Polosukhin. 2017.
\newblock Attention is all you need.
\newblock In \emph{NIPS '17}.

\bibitem[{Wang et~al.(2019)Wang, Su, Wang, Ji, and Ding}]{WSW+19}
Ran Wang, Haibo Su, Chunye Wang, Kailin Ji, and Jupeng Ding. 2019.
\newblock \href {http://arxiv.org/abs/1907.05338} {To tune or not to tune? how
  about the best of both worlds?}
\newblock \emph{CoRR}, abs/1907.05338.

\bibitem[{Wang and Jiang(2017)}]{WJ17}
Shuohang Wang and Jing Jiang. 2017.
\newblock \href {https://openreview.net/forum?id=HJTzHtqee} {A
  compare-aggregate model for matching text sequences}.
\newblock In \emph{5th International Conference on Learning Representations,
  {ICLR} 2017, Toulon, France, April 24-26, 2017, Conference Track
  Proceedings}. OpenReview.net.

\bibitem[{Xiong et~al.(2017)Xiong, Dai, Callan, Liu, and
  Power}]{DBLP:conf/sigir/XiongDCLP17}
C.~Xiong, Z.~Dai, J.~Callan, Z.~Liu, and R.~Power. 2017.
\newblock \href {https://doi.org/10.1145/3077136.3080809} {End-to-end neural
  ad-hoc ranking with kernel pooling}.
\newblock In \emph{Proceedings of the 40th International {ACM} {SIGIR}
  Conference on Research and Development in Information Retrieval, Shinjuku,
  Tokyo, Japan, August 7-11, 2017}, pages 55--64.

\bibitem[{Yang et~al.(2016)Yang, Ai, Guo, and
  Croft}]{Yang:2016:ARS:2983323.2983818}
L.~Yang, Q.~Ai, J.~Guo, and W.~B. Croft. 2016.
\newblock anmm: Ranking short answer texts with attention-based neural matching
  model.
\newblock In \emph{CIKM '16}.

\bibitem[{Yang et~al.(2019)Yang, Cer, Ahmad, Guo, Law, Constant, {\'{A}}brego,
  Yuan, Tar, Sung, Strope, and Kurzweil}]{YCA+19}
Y.~Yang, D.~Cer, A.~Ahmad, M.~Guo, J.~Law, N.~Constant, G.~Hern{\'{a}}ndez
  {\'{A}}brego, S.~Yuan, C.~Tar, Y.~Sung, B.~Strope, and R.~Kurzweil. 2019.
\newblock \href {http://arxiv.org/abs/1907.04307} {Multilingual universal
  sentence encoder for semantic retrieval}.
\newblock \emph{CoRR}, abs/1907.04307.

\end{thebibliography}

\clearpage
\appendix

\section{More Information On \ptdata}
\label{sec:app:ptdata}
We provide a few examples from the training data to better illustrate the goal of each task.
\paragraph{Product One}
\begin{itemize}
    \item Title: Aurora Dragon Fantasy Mink Blanket [Weight: Medium - 5LBS,Size: Queen].
    \item Description: Measures 79 inch x 96 inch and will fit a Queen of Full size bed.  Soft and plush.  Looks great and you will love cuddling up with at night.
    \item Sample terms and ratings
    \begin{itemize}
        \item size queen: 0.83
        \item dragon fantasy: 0.83
        \item size: 0.16
        \item T-shirt: 0.
    \end{itemize}
\end{itemize}

\paragraph{Product Two}
\begin{itemize}
    \item Title: Versace Women's Chain Reaction Chunky Sneakers - Size 37 (7).
    \item Description: The classic sneaker is given a haute update with experimental details-like a lightweight, chain-linked rubber sole and a riot of color and texture-for a must-have addition to your sneaker collection.  Style Name:Versace Chain Reaction Sneaker (Women).  Style Number: 5663881.
    \item Sample terms and ratings
    \begin{itemize}
        \item sneakers: 0.91
        \item leather: 0.08
        \item women: 0.58
        \item size 37: 0.78
    \end{itemize}
\end{itemize}

The ratings are aggregated from 3 human raters.

\section{Deriving \qpdata~ from MS Marco Ranking}
\label{sec:app:qpdata}
For pairwise input, creating a transfer set that roughly follows the same distribution as the training data can be very challenging (this is, however, not a problem in industrial systems as we can easily mine unlabeled data through logs). To this end, we utilize MSMARCO Passage Ranking data as it is of large scale, and we can easily create a large amount of unlabeled data. MSMARCO Passage Ranking is designed for ranking tasks, and it has 1M+ queries and 8.8M+ passages. Other popular datasets (e.g., GLUE benchmark) are relatively small, and previous distillation works often use text augmentation techniques to create transfer set.

In our work, we would like to \emph{directly} verify the effectiveness of model distillation, so instead of using ranking metrics (a decent scoring model does not always lead to better ranking metrics), we derive a binary classification task from the MSMARCO data,
\begin{itemize}
    \item First, all the human-rated query to passage pairs in MSMARCO Passage Ranking data are positive. We use that part as our positive examples.
    \item To create relatively hard negative pairs (so that the binary classification task can be more challenging), we encode queries/passages with the universal-sentence-encoder-qa\footnote{Available in https://tfhub.dev/google/universal-sentence-encoder-qa/3} \cite{YCA+19,chidambaram-etal-2019-learning} and run nearest neighborhood search (some public tools are available, e.g., \cite{JDH17}) to retrieve top-30 most relevant passages for each query. We then sample pairs with dot product below 0.53 as the negative pairs. The number of negative pairs is roughly the same as the number of positive pairs.
    \item For the transfer set, we simply retrieve the top-50 most relevant passages (measured via dot product of the query embedding and the passage embedding) and use those query/passage pairs as the unlabeled data.
\end{itemize}

\section{Poly-Encoders Fails for Long Text}
\label{sec:d2d}
\newcommand{\pp}{{\sc P2P-Rel}}
\newcommand{\poly}{{\sc PolyEncoders}}
Compared with DiPair, Poly-Encoders \cite{HSL+20} has at least the following limitations,
\begin{enumerate}[topsep=0pt, itemsep=-0.5em]
    \item It makes a strong assumption on its input pairs: One side of the input pair should be short text (e.g., less than 20 tokens).
    \item It does not extend to n-ary input.
    \item It can not deal with tasks beyond regression / binary-classification.
\end{enumerate}
Both 2. and 3. can be implied directly from the architecture of Poly-Encoders and assumption 1 is explicitly mentioned in \cite{HSL+20}. In this section, we experimentally show that when the assumption in 1. is violated, Poly-Encoders becomes considerably worse than DiPair.

We use an internal product to product similarity dataset (\pp). The average length of products is about 100, and Pearson correlation between model predictions and the human ratings is our primary metric. Our teacher model is a fine-tuned BERT-base model with a customized vocabulary, and our distillation set has 182M pairs.

\begin{table}[!h]
    \centering
    \small
    \begin{tabular}{|c|c|c|c|}
        \hline
        Model Settings & N & M & Pearson \\
        \hline
        Teacher & - & - & 0.840 \\
        \hline\hline
        \denmtsf & 6 & 6 & \textbf{0.826} \\
        \hline
        \poly & 1 & 11 & 0.805 \\
        \hline\hline
        \denmtsf & 3 & 3 & \textbf{0.823} \\
        \hline
        \poly & 1 & 5 & 0.790 \\
        \hline
    \end{tabular}
    \caption{\denmtsf~v.s. \poly~on \pp~data. We fix K=1. For fair comparison, we remove the projection layer in both methods as a projection layer is not proposed in Poly-Encoders.}
    \label{tb:d2d}
\end{table}

Consider the fact that a product has only about 100 tokens, we believe that for longer text such as full-page documents, the gap between \poly~and \denmtsf~will be even larger. We leave the verification of our hypothesis as future work.

\end{document}